\documentclass[conference]{IEEEtran}
\IEEEoverridecommandlockouts
\usepackage[sorting=none]{biblatex} %Imports biblatex package
\usepackage{amsmath,amssymb,amsfonts}
\usepackage{algorithmic}
\usepackage{graphicx}
\usepackage{textcomp}
\usepackage{xcolor}
\usepackage{authblk}
\usepackage{float}
\usepackage{tabularx}
\usepackage{hyperref}
\usepackage{tikz}
\usepackage{lipsum}
\usepackage{float} %picture formatting 
\renewcommand{\baselinestretch}{0.95}
\newcommand\copyrighttext{%
  \footnotesize \copyright 2024 IEEE. Personal use of this material is permitted. Permission from IEEE must be
obtained for all other uses, in any current or future media, including
reprinting/republishing this material for advertising or promotional purposes, creating new
collective works, for resale or redistribution to servers or lists, or reuse of any copyrighted
component of this work in other works.}
\newcommand\copyrightnotice{%
\begin{tikzpicture}[remember picture,overlay]
\node[anchor=south,yshift=10pt] at (current page.south) {\fbox{\parbox{\dimexpr\textwidth-\fboxsep-\fboxrule\relax}{\copyrighttext}}};
\end{tikzpicture}%
}

\def\BibTeX{{\rm B\kern-.05em{\sc i\kern-.025em b}\kern-.08em
    T\kern-.1667em\lower.7ex\hbox{E}\kern-.125emX}}

\begin{document}

\title{A proximal policy optimization based intelligent home solar management \\}

\author{Kode Creer  }
\author {Imtiaz Parvez}
\affil{Department of Computer Science, Utah Valley University, 800 W University Pkwy, Orem, UT 84058}
\affil{{Email:\{{kode.creer, imtiaz.parvez\}@uvu.edu}}}

\maketitle

\begin{abstract}
In the smart grid, the prosumers can sell unused electricity back to the power grid, assuming the prosumers own renewable energy sources and storage units. The maximizing of their profits under a dynamic electricity market is a problem that requires intelligent planning. To address this, we propose a framework based on Proximal Policy Optimization (PPO) using recurrent rewards. By using the information about the rewards modeled effectively with PPO to maximize our objective, we were able to get over 30\% improvement over the other naive algorithms in accumulating total profits. This shows promise in getting reinforcement learning algorithms to perform tasks required to plan their actions in complex domains like financial markets. We also introduce a novel method for embedding longs based on soliton waves that outperformed normal embedding in our use case with random floating point data augmentation. 

\end{abstract}


\copyrightnotice

\section{Introduction}
Consumers’ active participation in the energy management process is an essential characteristic of envisioned future smart grid systems. Such prosumers consume energy but also own energy production units (e.g. solar panels) and storage units; they can make decisions to buy and sell the energy they produce based on market information. It is anticipated that a significant amount of produced energy will come from green sources and prosumer-possessed infrastructure. Prosumer’s active participation changes the dynamics of how energy is produced, priced, and consumed. In the prosumer-centric future smart grid, the aggregate effect of end-user decisions will impact global policies for grid management and outcomes such as efficient generation and distribution. One challenge faced by prosumers is selecting an active energy trading strategy (or policy) in the face of dynamic price changes. 

In the smart home grid, prosumers may possess various renewable energy sources such as solar panels and windmills along with storage that stores extra energy. The prosumer can either hold or sell the energy back to the power grid. There are fluctuating prices and unpredictable amounts of how much energy will come in. Maximizing the accumulation of the metric also is not a Markov decision process problem. Therefore this presents additional challenges. This calls for an intelligent management system/agent that prudently manages the produced solar energy. 

The goal of the intelligent agent is to manage solar energy and sell it based on the observation feed into a model. In this regard, we use Proximal Policy Optimization (PPO) and Sparse Mixture of Experts (MOE) for our time series forecasting algorithm for testing.  Sparse Mixture of Experts is a method we use in our time series forecasting to allow larger models with less overfitting to get better generalizations on less data~\cite{DBLP:journals/corr/ShazeerMMDLHD17}. PPO is a training method for reinforcement learning that is trained for maximizing objectives in complex environments. PPO has shown to  achieve state of the art in many tasks~ \cite{DBLP:journals/corr/SchulmanWDRK17}.

Our task is to perform smart home solar energy management. This problem requires the model to be both directly or indirectly capable of forecasting prices, and able to predict the most optimal time to accumulate rewards over time.  In the literature, few works can be found that focus on cost minimization between the solar energy generated and the electricity demand from the different kinds of appliances~ \cite{10066193,} \cite{HUY2023117340,}. Lee et al's work combines both a supervised model to predict key variables like temperature to optimize the policy instead of just a simple AI agent~\cite{s19183937,}. This comes with the benefit of providing extra information to the AI agent.  There are also other approaches that use optimization engines instead of reinforcement learning for management that show promising results over conventional methods~\cite{6599649,}. There are other methods that show that applying game theory can help reduce the uncertainty of price changes in a dynamic market~\cite{8031035,}  and minimize the costs of energy utilization~\cite{6835827, 5628271,}.  Unlike these similar works, our problem is assuming no power usage is counted into the mix and it doesn't focus on the distribution of energy to appliances. Our goal is to explicitly maximize the total accumulated profits from selling energy back to the grid. We feel that focusing on this allows us to apply our findings to other domains like natural language processing and managing financial markets. 

Most reinforcement algorithms like PPO are designed for a Markov decision process, which means that each state must not be reliant on the previous states. 
The Markov decision process with no adjustments presents limitations of a problem that does rely on previous states. There are methods to try to get around this like this work's attempt to using Monte Carlo Tree Search policy trained using PPO \cite{liu2023dont,}. The strength behind this approach is that it is able to plan ahead its actions and facilitate exploration in a stable way. This also gets more out of the decision process in the value estimation function. The main problem with using Monte Carlo Tree Search is the increasing memory requirements of storing each state in the tree. For every day there are two possible actions buying and selling. For every day there are two actions holding or selling. Therefore for each day $n$, there are \(2^n\), where $n$ is days, possibilities. This makes it not feasible in our use case to use this method. Other methods try to optimize by using behavior cloning through reinforcement learning \cite{aruoss2024grandmasterlevel,} to get around the Markov decision process, but our method does not rely on behavior cloning to find optimal rewards.

The contribution of our work are as follows:
\begin{enumerate}
    \item A way to automate and improve the energy management of solar power on the residential premise using PPO designed to handle low amounts of data.
    \item A method for structuring rewards and sparse rewards that allows PPO to improve its performance in long-term contexts
    \item A method for data augmentation and an embedding based on soliton that outperformed Embeddings for our use case on limited data and using a Sparse Mixture of Experts model. 
\end{enumerate}

The paper is structured as follows. Section II is the methodology that also cover how we implemented the algorithms used in our experiments. Section III goes over the results and simulation. This  contains our data and results. Section IV is the conclusion that summarizes our closing remarks from our experiment. 

\section{Methodology}

We created an offline reinforcement learning environment and separated our data into a training dataset and a test environment. The agent is given a fixed window to liquidate. In our case, it's the end of the test environment. This is not an optimal strategy, but it's a method to minimize losses from not selling. The data we are using is trained on data from the 2016 NREL and US Energy Information.
We adjust the data from the work of Parvez et al~\cite{10115183} to make the environment less noisy. We combined the power generated into a single observation instead of breaking it down into 48 timesteps per day. Each AI agent will be tested on for 30 episodes in the test environment. We only test sell only for one episode and random for 5 episodes. This is because there is less of a need due to the deterministic nature of the actions being taken. 

We calculate the amount of energy by using the VMP and IMP and using the prices given from Mid C Peak to calculate the returns. 

Wattage is used for us to help calculate the balances, but the data only provides metrics to calculate them.  
Wattage is calculated as follows
\begin{equation}
Watt = v_{mp}\cdot i_{mp}
\end{equation}

We will compare the performance of our PPO agent to an algorithm that performs sell-only, random choices, and MOE time series forecasting with an algorithm that picks the best day to sell.

We choose to do these methods  for the following reasons
\begin{enumerate}
    \item Sell only has us able to tell if the agent is able to plan ahead or not. It helps us make sure the agent isn't trying to maximize each time step instead of the total sum of rewards. 
    \item Random and time series forecasting is to make sure that the PPO agent is making valid choices and not making lucky guesses when it does sell. 
    \item All of them give solid indicators of the kind of actions the agent is making under the hood. 

\end{enumerate}

For MOE, we train it on just the daily prices into its own random split of a test and training set. We then use an algorithm to pick the best day to sell from its time series forecasting. There is however a chance it views prices in days that are seen in the test environment because it has its own independent random split for testing data. Our time series forecasting only trained on predicting the prices.

These are the following algorithms used in their experiments and how we implemented them below. 

\subsection{Proximal Policy Optimization (PPO)}

PPO is a training method in reinforcement learning. The objective of PPO is modeled as follows. 
\begin{equation}
\begin{split}
\text{PPO Objective} = \mathbb{E}\Biggl[&\min\Biggl(\frac{\pi(a|s)}{\pi_{\text{old}}(a|s)}\hat{A}(s, a), \\
&\text{clip}\left(\frac{\pi(a|s)}{\pi_{\text{old}}(a|s)}, 1 - \epsilon, 1 + \epsilon\right)\hat{A}(s, a)\Biggr)\Biggr]
\end{split}
\end{equation}

where $a$ and $s$ stand for respectively. $\hat{A}$stands for the actor model. 
This is what each of the variables stands for, $\pi$ stands for the policy

This is usually accompanied with a advantage estimation function, but not always. We also include a fixed window for the agent to sell during inference. In our case its just the end of the episode. This is far from the optimal policy, but it provides a fallback mechanism to minimize the losses from the algorithm taking risky actions.

\subsection{Generalized Advantage Estimation ( GAE)}

We will use Generalized Advantage Estimation ( GAE)\cite{schonlau_random_2020} for our advantage estimation function. In order to speed up training we attempted to vectored this and was able to get quicker training times.
Our advantage function is modeled as follows.
\begin{equation}
\text{General Advantage Estimate (GAE)} = \sum_{t=0}^{T-1} \left(\delta_t + (\gamma\lambda)\delta_{t+1}\right)
\end{equation}

$\delta$ stands for the delta at time step $t$. $\gamma$ and $\lambda$ are represented by their respective characters and the delta at time step $t+1$. \\

\subsection{Soliton embedding}\label{BB}
The soliton embedding is a based on Soliton waves due to their ability to represent non-linear dynamics. This is based on the Integration of the $KdV$ equation for finding the shape of the soliton 

\[
A \sec^{2}[b(x-vt)]
\]

where $A$ is free parameter , $b$ is scaling factor, $v$ is the velocity, and $t$ is the time. 

However, it's with the following adjustments. We use a MLP to predict $A$ value with bias. We also have another MLP that is able to predict $b(x-vt)$. The full equation is modeled as follows.
\[
bxvt = Tanh(MLP(x))
\]
\[
A = MLP(x)
\]
\[
A * (1/ Cos(bxvt))^2 + Sin(bxvt/Cos(bxvt))
\]
$x$ stands for the input to be passed into our embedding. 

To take advantage of the inputs that are modeled by longs. We also add a random float between 0 and 0.99 to add to the input value to add more diversity to the data and better interpolation between values. 

\subsection{Sparse Mixture of experts}\label{BB}
For the supervised model, we are using a sparse mixture of experts, training it to forecast the price according to the day of the year. In the environment, we have it forecast the prices in a time window and have it sell on the day with the highest price. This is how the algorithm picks the best day to sell. 

\subsection{Reccurent Rewards}\label{BB}
We put the reward function in the observation and include the previous and current reward functions as the reward. This allows it to contextualize the importance of each action in a longer-term view compared to a simple Markov function that focuses on maximizing the profit at each time step. Our observation is modeled as follows. 
\[
   State = [p, w, pg, r]
\]

$p$ stands for the price of the energy. $w$ stands for the wattage in the power storage. $pg$ stands for the power generated. $r$ stands for the reward. 

By putting the rewards into the state, it allows the model to change its actions based on the previous state similar to how a recurrent neural network like LSTM stores previous information in its hidden state\cite{10.1162/neco.1997.9.8.1735,}. 

The way the rewards were calculated was just as important as including it as an observation value. Rewards for each time step were calculated as follows.
\begin{equation}
\begin{split}
\text{reward} & = \text{Max(0, last reward} - \\
 & \text{Voltage generated} + \text{current balance})
\end{split}
\end{equation}

We found that adding the balance improved results in our testing environment and having the reward clamped at a minimum of 0 helped stability during training. 

In addition to the previous reward function getting passed into the observation, we also make the rewards sparse and train it from data when it sells. Therefore optimizing when to sell in a given state.

For the model architecture, we use a simple 4-layered mlp for the actor and a 3-layered mlp for the critic. We avoid using context-aware models like RNNs, Transformers, etc... to ensure that our recurrent rewards framework is the reason for the results rather than the neural architecture itself.

A combination of these two with the fixed window to sell is crucial to our policy.

\section{Results and Simulation}

Our PPO agent was able to get superior results over the algorithms by at least by around 30\%. 

Table~I demonstrates the results of each of the agents we have. The trend is the longer we train the PPO agent, the higher the total rewards. Column 1 is the performance of the MOE forecasting algorithm. Column 2 is the performance of the AI agent when trained on 30 epochs and Column 3 is the same agent when trained on 1000 epochs.
\begin{table}[H]
\centering
\caption{MOE, PPO 30 epochs, and PPO 1000 epochs}
\label{tab:my-table}
\begin{tabular}{|c|c|c|}
\hline
MOE  & PPO 30 epochs & PPO 1000 epochs \\ \hline
14.99 & 54.68 & 124.46 \\ \hline
15.03 & 33.34 & 33.34 \\ \hline
15.13 & 33.34 & 33.34 \\ \hline
14.99 & 9.69 & 33.34 \\ \hline
14.58 & 13.6 & 33.34 \\ \hline
14.98 & 15.01 & 33.34 \\ \hline
13.84 & 10.76 & 10.64 \\ \hline
14.66 & 14.14 & 13.74 \\ \hline
15.01 & 13.58 & 16.36 \\ \hline
14.46 & 13.58 & 16.41 \\ \hline
14.97 & 12.4 & 13.38 \\ \hline
14.81 & 15.86 & 14.48 \\ \hline
14.93 & 10.62 & 15.46 \\ \hline
13.85 & 10.77 & 15.46 \\ \hline
15.2 & 9.8 & 15.46 \\ \hline
13.88 & 9.01 & 15.46 \\ \hline
13.8 & 9.41 & 15.46 \\ \hline
14.55 & 11.72 & 15.46 \\ \hline
15 & 10.44 & 16.57 \\ \hline
13.79 & 12.46 & 16.93 \\ \hline
14.52 & 12.28 & 12.5 \\ \hline
13.88 & 14.57 & 15.4 \\ \hline
14.51 & 24.67 & 15.4 \\ \hline
15.1 & 25.26 & 16.65 \\ \hline
14.9 & 25.86 & 17.22 \\ \hline
14.47 & 11.78 & 10.89 \\ \hline
14.76 & 13.66 & 13.11 \\ \hline
13.86 & 13.66 & 13.11 \\ \hline
14.51 & 10.38 & 13.33 \\ \hline
15.15 & 13.75 & 18.33 \\ \hline
\end{tabular}
\end{table}
Sell Only has a value of 14.5. We got this value from running this strategy through the test environment. Because it runs the same actions in the same test environment, it will consistently yield the same result.

Table~II demonstrates  the performance of the random selection in the same test environment.
We only ran Random numbers for 5 trials. 

\begin{table}[H]
\centering
\caption{Random selection performance}
\label{tab:random-table}
\begin{tabular}{|c|}
\hline
Random \\ \hline
18 \\ \hline
15.05 \\ \hline
15.09 \\ \hline
14.55 \\ \hline
15.06 \\ \hline
\end{tabular}

\end{table}

Figure 1 shows the averages across each trial from the data across the columns in Table 1.
\begin{figure}[H]
    \centering
    \caption{Agent Performance}
    \label{fig:enter-label}
    \includegraphics[width=1.0\linewidth]{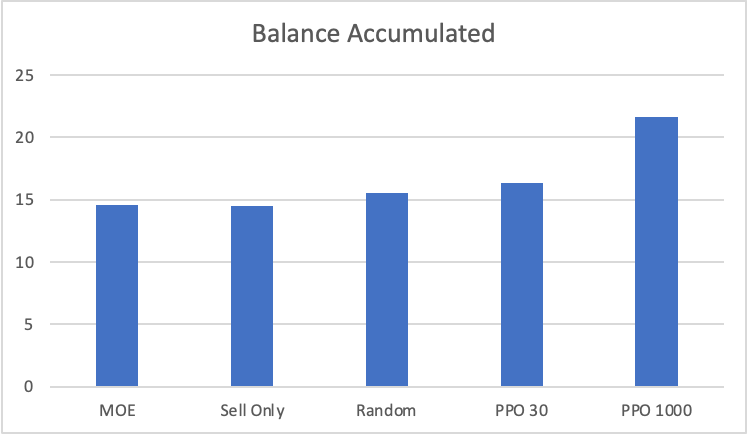}
    
\end{figure}

Table~III shows the exact numbers of each of the values in Figure 1.
\begin{table}[H]
\centering
\caption{Data Table}
\label{tab:data-table}
\begin{tabular}{|c|c|c|c|c|}
\hline
MOE & Sell Only & Random & PPO 30 & PPO 1000 \\ \hline
14.60367 & 14.5 & 15.55 & 16.336 & 21.61233 \\ \hline
\end{tabular}

\end{table}

Figure 2 demonstrates the relationship between the epochs trained and performance over epochs trained. The trend shows minor increases in performance as we have the PPO agent train for more episodes.
\begin{figure}[H]
    \centering
    \caption{Performance of PPO agent over episodes trained}
    \label{fig:enter-label}
    \includegraphics[width=1\linewidth]{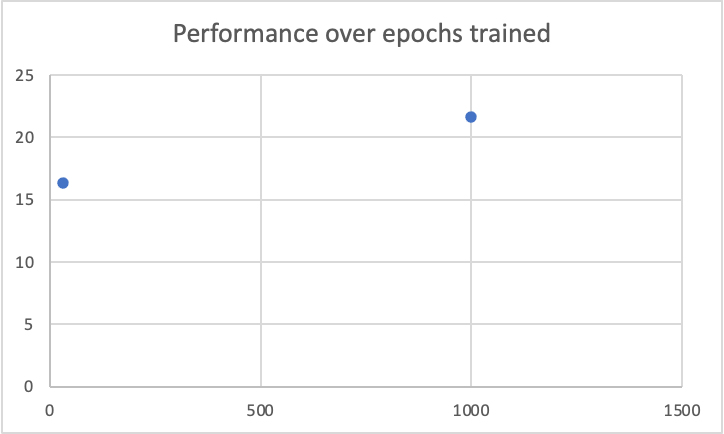}
    
\end{figure}

PPO was able to beat sell only by a noticeable increase in performance despite the low amount of data. We had to give it a 30\% split for the validation data to give it enough time to reap the efforts of long-term actions.

As far as the results of the MOE on supervised training. It has its own split for validation of the prices. There is probably exposure to prices in the evaluation set in the environment.  We first tried to run the prices through an embedding. It has a great ability to overfit but failed to generalize at a satisfactory quality on its testing data. So we tried an embedding based on soliton waves. Without major tuning, it was able to match the embedding performance and was much less prone to overfitting than the embedding. This allowed us to put in more parameters without sacrificing the consequences of overfitting. \\

Figure 3 shows the results and the tuned parameters of each trial.  This goes over the training Mean Squared Error with the normal embedding to model the relationship between the number of experts and performance achieved on the training dataset.

\begin{figure}[H]
    
    \centering
    \caption{Training results on normal embedding}
    \label{fig:enter-label}
    \includegraphics[width=1\linewidth]{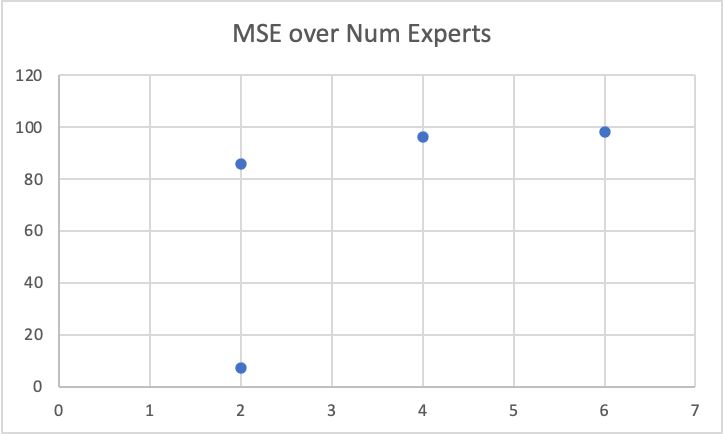}
    
\end{figure}

Table IV demonstrates the performance of the embeddings. Column 1 refers to the dimensions of the embedding and dimensions of the number of hidden layers per expert if mentioned, Column 2 has the number of experts, and Columns 3 and 4 have the train and test metrics. 
\begin{table}[H]
\centering
\caption{Normal Embedding}
\label{tab:normal-embedding}
\begin{tabular}{|c|c|c|c|}
\hline
Dimensions & Experts & Train MSE Loss & Test RMSE Loss \\ \hline
128  & 2 & 7.3115 & 9.6131 \\ \hline
128 (64 dim per expert) & 2 & 85.9594 & 9.0421 \\ \hline
128 (64 dim per expert) & 4 & 96.0857 & 8.3847 \\ \hline
128 (64 dim per expert) & 6 & 98.2492 & 10.7752 \\ \hline
\end{tabular}

\end{table}

Table~V shows the results of the soliton embedding. Column 1 refers if hyperparameter optimization happened  to the model or not and the number of experts. Optimal refers to our model with manual hyperparameter tuning.
\begin{table}[H]
\centering
\caption{Soliton Embedding and Float Data Augmentation}
\label{tab:float-augmentation}
\begin{tabular}{|c|c|c|}
\hline
Tuning & Train MSE Loss & Test RMSE Loss \\ \hline
No (6 experts) & 115.0481 & 8.1353 \\ \hline
Yes (6 experts, optimal) & 53.7173 & 5.4411 \\ \hline
\end{tabular}

\end{table}

Figure 4 shows the results of the table above displayed as a bar chart to compare the performance of each. The lower the values in orange, the better it is able to generalize on the test set. This also shows how generalized each model is comparing test and train losses. 
\begin{figure}[H]
    \centering
    \includegraphics[width=1\linewidth]{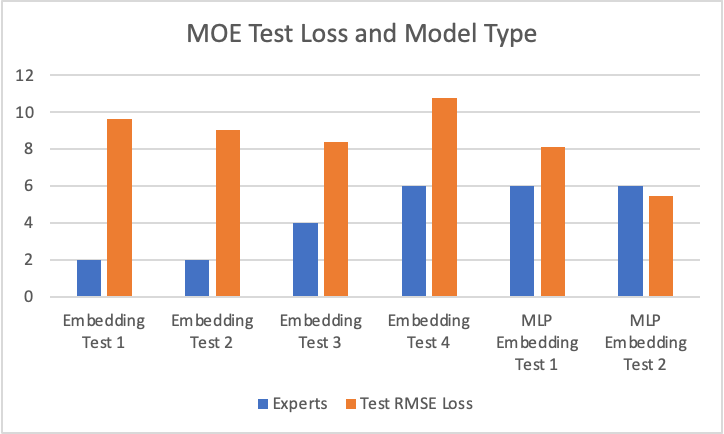}
    \caption{Test performance on embedding. Organized left to right in same order as tables}
    \label{fig:enter-label}
\end{figure}

\section {Conclusion}
In conclusion, we were able to get superior performance for trading smart home solar energy using PPO with adjustments to the reward function to carry the previous reward into its current reward. Combining this with sparse rewards and a fixed time window to liquidate allows it to get higher gains on average compared to other naive algorithms on such minimal data. We expect with more data and better neural network architectures that there would be less of a need for the fixed window to sell in low-risk tasks. We also found augmenting data using longs by adding random decimals between 0 and 0.99 and putting it through a soliton embedding that can take in both floats and integers improved generalization. This can lead to future work looking into new ways of augmenting data for language models without ruining the meaning behind the text.  Our findings also provide a systematic way to model rewards that have a desired objective that relies on previous steps without a tree search.

\section*{Acknowledgment}

This work was supported by the Utah Valley University Undergraduate Research Scholarly and Creative Activities Grant. 

We open source our implementation in this git hub repo to facilitate additional research.
\href{https://github.com/kodecreer/SolorEnergyManagerAI}{repository link: [https://github.com/kodecreer/SolorEnergyManagerAI]}

\printbibliography
% \bibliography{biblio}
% \bibliographystyle{IEEEtran}

\end{document}